\documentclass[a4paper,conference]{IEEEtran}

\usepackage{caption,newfloat}
\DeclareCaptionType{Model}

\ifCLASSINFOpdf
  \usepackage[pdftex]{graphicx}
  \graphicspath{{../images/}}
  \DeclareGraphicsExtensions{.pdf,.jpeg,.png}
\else
  \usepackage[dvips]{graphicx}
  \graphicspath{{../eps/}}
  \DeclareGraphicsExtensions{.eps}
\fi
\usepackage{amsmath}
\usepackage{algorithmic}
\usepackage{array}
\ifCLASSOPTIONcompsoc
  \usepackage[caption=false,font=normalsize,labelfont=sf,textfont=sf]{subfig}
\else
  \usepackage[caption=false,font=footnotesize]{subfig}
\fi

\usepackage{stfloats}
\usepackage{url}
\begin{document}
\title{Cross-Subject Deep Transfer Models for Evoked Potentials in Brain-Computer Interface}

% \author{\IEEEauthorblockN{Chad Mello\\ and Troy Weingart}
% \IEEEauthorblockA{Department of Computer \& Cyber Sciences\\
% United States Air Force\\
% Colorado Springs, CO 80840\\
% Email: chad.mello@afacademy.af.edu, troy.weingart@afacademy.af.edu}
% \and
% \IEEEauthorblockN{Ethan Rudd\\}
% \IEEEauthorblockA{OmniScience LLC\\
% 	Colorado Springs, CO\\
% 	Email: ethan@omnisciencellc.com}}

\author{\IEEEauthorblockN{Chad Mello\IEEEauthorrefmark{1},
Troy Weingart\IEEEauthorrefmark{2}, and Ethan M. Rudd\IEEEauthorrefmark{3} }
\IEEEauthorblockA{\IEEEauthorrefmark{1}\IEEEauthorrefmark{2} U.S. Air Force Academy
\IEEEauthorrefmark{1}\IEEEauthorrefmark{3} OmniScience LLC\\
%Wherever\\
Email: \IEEEauthorrefmark{1}\IEEEauthorrefmark{2}\{chad.mello,troy.weingart\}@afacademy.af.edu,
\IEEEauthorrefmark{3}ethan@omnisciencellc.com}
}

\maketitle

% \begin{abstract}
% Brain-Computer Interface (BCI) technology can improve the lives of millions of people around the world, be it through assistive technologies or clinical diagnostic tools.  Despite the overwhelming evidence demonstrating immense interest in BCI across a multitude of domains, along with advancements in hardware technologies that supports it, consumer and clinical viability remains low.  We address some of the predominant challenges that deter widespread availability of reliable BCI, demonstrating ways to augment system bootstrapping with pretrained deep classifiers that perform well out of the box against raw electroencephalographic (EEG) data for cross-subject tasks.

% We repurpose an older, but well-curated EEG dataset collected for the original purpose of clinical research and turn it into a transfer learning benchmark.  We compare our novel convolutional model benchmarks against a variety of common EEG feature-based
% learning approaches and techniques, providing evidence that our deep transfer learning approach performs better overall
% without explicitly requiring localized features over evoked potentials.  Indeed, our experiments lead us to believe that transfer learning may assist in the development of new classifiers that offer superior performance over classifiers currently applied to BCI.
% \end{abstract}

\begin{abstract}
Brain Computer Interface (BCI) technologies have the potential to improve the lives of millions of people around the world, whether through assistive technologies or clinical diagnostic tools. Despite advancements in the field, however, at present consumer and clinical viability remains low. A key reason for this is that many of the existing BCI deployments require substantial data collection per end-user, which can be cumbersome, tedious, and error-prone to collect. We address this challenge via a deep learning model, which, when trained across sufficient data from multiple subjects, offers reasonable performance out-of-the-box, and can be customized to novel subjects via a transfer learning process. We demonstrate the fundamental viability of our approach by repurposing an older but well-curated electroencephalography (EEG) dataset and benchmarking against several common approaches/techniques. We then partition this dataset into a transfer learning benchmark and demonstrate that our approach significantly reduces data collection burden per-subject. This suggests that our model and methodology may yield improvements to BCI technologies and enhance their consumer/clinical viability.
\end{abstract}

% \fi
\IEEEpeerreviewmaketitle

\section{Introduction}
\label{sec:intro}

% Things to mention here...

% 1. Traditional BCI has involved very problem specific approaches, features, data collection, etc.

% 2. Notable exceptions discussed in background but they do not generalize 

% 3. Our approach utilizes combinations of:
%	a) learning from raw data 
%	b) transfer learning 
%	c) interpretability, inspired from vision domains

 BCI captures, measures and records brain function via brainwaves and relates those measurements to some sort of external classification, action, or task~\cite{wolpaw2012brain, EEG_Signal_Processing, QuantitativeEEG}.  These recordings, typically captured via electrodes placed externally over the scalp, are referred to as \textit{electroencephalography} (EEG). EEG contains \textit{event-related potentials} (ERP) -- measured brain responses that are generated by the brain's cognitive, sensory, or motor functions.  ERP may be time-locked with both sensory and cognitive neuronal processes and then turned into executable or labeled actions by the BCI.  BCI systems rely on classifiers designed to identify and classify ERP related to tasks within the context of a specific system design.

At the time of this writing consumer and clinical BCI viability remains low, due to problematic aspects associated with EEG data: EEG is noisy, non-linear and highly non-stationary~\cite{EEG_Signal_Processing, QuantitativeEEG}, and ERP varies spatially and temporally between different mental tasks as well as across users~\cite{chiou2016spatial, li2020transfer}. Moreover, collection can be time-consuming, tedious, and error-prone sometimes leading to a dirth of available data in comparison to other machine learning domains, e.g., computer vision datasets often consist of millions to hundreds of millions of images, whereas even large EEG datasets consist of only a few thousand trials. 
% citation for cv dataset sizes vs eeg dataset sizes

In consumer/clinical application, the BCI community tends to address these challenges via manually crafted features constructed with subject matter expertise. % citation on SME features used for consumer/clinical BCI applications
This featurization approach makes extracting relevant signal and fitting classifiers feasible on relatively small datasets for a specific subject and a specific task. However, such features also introduce a high degree of inductive bias~\cite{goyal2020inductive} which may attenuate relevant responses associated with other tasks, including cross-subject correlation. Our approach utilizes a deep learning architecture, which operates directly on filtered EEG signals, yet is simultaneously engineered for fitting on EEG-scale datasets. Our approach is thus application-agnostic. 

We hypothesize that our approach 1) can learn a representation which works well across subjects in generic form and 2) can consequently be domain-adapted to novel subjects with a specific BCI application under a transfer learning regime. This work makes the following contributions:

\begin {itemize}
\itemsep-0.25em

\item Novel evaluation of transfer learning approaches towards generalization techniques that allow for simplified and shortened classifier training specific to EEG. 

\item Evaluation of traditional EP BCI classifiers with respect to performance on filtered EEG data (i.e. no hand-crafted features). 

\item Evaluation of EP classifiers with respect to generalization potential across subjects and tasks.
\end {itemize}

\section{Background}
\label{sec:background}

BCI are built around one or more types of ERP buried in EEG, and when matched to tasks (i.e. a given context), these ERPs provide the basic building blocks for a BCI system. There are a number of challenges intrinsic to EEG data, for example, filtered (e.g., via a bandpass filter) EEG data is high-dimensional, has a low signal-to-noise ratio (SNR), and is \textit{highly} non-stationary. Evoked potentials (EPs) in EEG have an explicit dependence on the time variable (an EP is time-locked with the stimulus), making more input from the past necessary for classification and regression models. EEG contains biological artifacts such as eye blinks, muscle artifacts, and head movements~\cite{QuantitativeEEG}. 

Many of what we refer to as \textit{traditional} classifiers were applied to BCI before research focus shifted to deep learning models.  Many of these classifiers are based on liner models, such as variations of linear discriminant analysis (LDA), principal component analysis (PCA), and support vector machines (SVMs) -- including non-linear kernel SVMs~\cite{aloise2012comparison, krusienski2006comparison}. Bayesian approaches as well as clustering techniques, ensemble learning such as random forests (RF) and gradient boosting (GB), hidden Markov models (HMMs), and distance metrics are also considered \cite{lotte2007review, akram2015efficient, mainsah2015increasing, verschore2012dynamic, helmy2008p300, speier2014integrating, yger2016Riemannian}.  Most of these approaches include some degree of feature engineering.  We use some of these classifiers in our experimental comparisons; however, our work is particularly focused on deep transfer learning models.

\subsection{Deep Learning in BCI}

Deep learning (DL) both has and continues to achieve state-of-the-art performance in many areas of applied machine learning \cite{lecun2015deep,hestness2019beyond}, and consequently, recent BCI research leverages deep learning as well. Deep learning has the advantage that it can learn directly from filtered EEG data, rather than (necessarily) relying on SME-derived features.  Moreover, DL models are less sensitive to localization than traditional ML models \cite{lecun2015deep}. 

However, training DL architectures on direct EEG data (as opposed to featurized data) requires additional labeled data and model regularization to avoid over-fitting \cite{hestness2019beyond}. This is an inherent limitation on performance achievable by current DL BCI classifiers, as even the relatively large EEG dataset that we use in this paper consists ``merely" of thousands of trials from 122 subjects. For perspective, the Celeb-500k face recognition dataset \cite{cao2018celeb} consists of 50 million images from 500k individuals, which, according to Hestess et al.~\cite{hestness2019beyond} is still orders of magnitude too small to fully leverage DL.  However, as EEG sensors become more ubiquitous, cost effective, and ergonomic, we anticipate that labeled EEG dataset sizes will increase and performance of DL BCI models will ultimately improve as a result. 

There are a host of papers presenting various neural network and deep network architectures and applications towards improving BCI performance \cite{vavreka2017stacked, maddula2017deep, abdulhay2017investigation, liu2016time, soltanpour2016enhance, achanccaray2017p300, zafar2017electroencephalogram,lawhern2018eegnet}. 

\subsection{Transfer Learning in BCI}

Overwhelming evidence demonstrates that transfer learning techniques can increase performance over a multitude of tasks for almost any deep learning domain~\cite{tan2018survey}. As BCI has adopted deep learning a number of transfer learning techniques, including deep transfer learning techniques, have been pioneered \cite{giles2019subject, uran2019applying, ozdenizci2019transfer, wang2015review, atyabi2017reducing, wu2014transfer, rodrigues2019exploring, zanini2017transfer, zheng2016personalizing, he2017transfer, chai2017fast}. These works have tackled challenging aspects of transfer learning that are unique to EEG data, including varying electrodes, amplifiers, and stimulus modalities. In contrast to these, our work focuses on training a common base DL model in a generic multi-subject regime and adapting the model to novel subjects in a way that decreases per-subject data collection burden; we seek to generalize across filtered EEG without dependencies on specific feature engineering, stimulus modality, EEG hardware, tasks, or subjects.  
\section{Approach}\label{sec:approach}

In Sec. \ref{sec:dataset} we describe the dataset and data splits that we use under two evaluation scenarios: performance benchmarking (Scenario 1) and transfer learning evaluation (Scenario 2). We then describe our DL architecture and elaborate on our design rationale (Sec. \ref{sec:dl_model}). Next, we introduce traditional models against which we benchmark, under Scenario 1 to establish the viability of our DL architecture (Sec. \ref{sec:benchmark_models}). Finally, in Sec. \ref{sec:transfer_benchmark}, we discuss different transfer models which we evaluate under Scenario 2. 

\subsection{Evaluation Scenarios and Dataset}
\label{sec:dataset}

For our experiments, we re-partitioned an old but well-curated dataset \cite{ingber1997statistical,ingber1998statistical}. The original dataset includes 122 subjects from two groups: the first group contains 77 subjects with a high risk of alcoholic predisposition (``High-Risk") while the second group is a control group of 45 subjects with low risk of  alcoholism (``Low-Risk").  To elicit event-related potentials (ERPs), a delayed matching-to-sample task was used. For each subject, trials were conducted in which two images appeared in succession with a 1.6 second inter-stimulus interval. First, a random image (S1) from the set was displayed to the subject for 300ms. The image that followed was either a match to the first (S2 Match), or a different image altogether (S2 No Match). For each trial, 1 second of EEG was collected at a 256 Hz sampling rate. Trials were then filtered using a bandpass filter. In our experiments, we use these filtered trials as is, with no re-sampling. While the images were shown, EEG measurements were collected using 64 scalp electrodes (channels) with a standard 10/20 extended spatial configuration \cite{homan1987cerebral}. Of the 64 channels, four serve as reference and/or ground, so we conducted our experiments using only the 60 non-reference channels or subsets thereof according to the same 10/20 configuration. Match/Non-Match trials were balanced, with 120 trials per subject. All trials were randomized and balanced; one half of the trials contained matches, while the other half did not. 
%where recordings were segmented into 1-second windows which included 3.9ms epochs.

For our experiments, we utilize only the EEG taken from the query image (S2 Match/No-Match), labeling each EEG trial according to ``Match/Non-Match" and ``High-Risk/Low-Risk" respectively. We re-partitioned the dataset from ~\cite{ingber1997statistical} and \cite{ingber1998statistical} according to two experimental scenarios. Scenario 1 is used to benchmark our deep learning architecture against traditional classifiers. Scenario 2 is used to evaluate different transfer learning approaches using our deep learning architecture.

\begin{figure}[h]
	%\vspace{-.70cm}
\centering
\includegraphics[scale=.25]{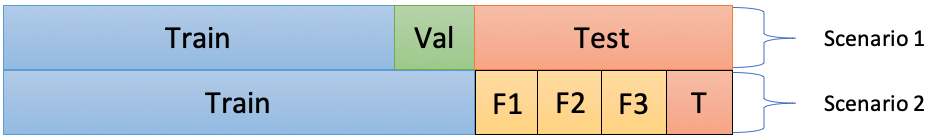}
\caption{\label{fig:partition} Dataset splits for Scenarios 1 (top) and 2 (bottom).}
% \caption{\label{fig:partition} Data partitions used for our original benchmarking (middle row) and transfer learning experiments (bottom row).  F1, F2, and F3 (up to \%75 of the remaining data) are sections of data used for fine tuning, leaving section T (25\%) for testing.}
\end{figure} 

%\subsubsection{Scenario 1: Performance Benchmarking}
\noindent
{\bf Scenario 1 -- Performance Benchmarking:}
We stratify our train, test, and validation partitions according to subjects, i.e., each partition contains all trials from selected subjects and no subject overlap exists between partitions. Subjects were sampled without replacement according to a uniform random distribution. Of the subjects, 70\% were assigned to the train set, 20\% were assigned to the test set, and 10\% were assigned to the validation set, which was used to establish hyperparameters.

\noindent
{\bf Scenario 2 -- Transfer Learning Evaluation:} In Scenario 2, we combine the training and validation partitions from Scenario 1 into one monolithic partition for pre-training transfer learning variants of our deep learning architecture. We then split our test partition from Scenario 1 into four novel equal-sized partitions: three fine-tuning partitions (F1,F2,F3) and a single test partition (T). Note that these splits were stratified over trials, sampled according to a uniform random distribution, meaning that F1, F2, F3, and T have subject overlap with respect to one another, but contain no subject overlap with respect to the pre-training partition.  

\begin{figure*}[!b]
        \centering
        \subfloat[High-Risk/Low-Risk]{\label{fig:alcoholicFull}
        \includegraphics[width=0.45\linewidth]{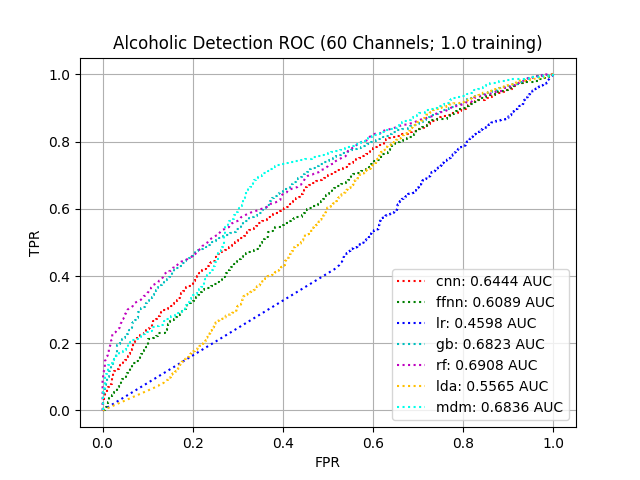}}
        \hfill
        \subfloat[Match/Non-Match]{\label{fig:matchFull}
        \includegraphics[width=0.45\linewidth]{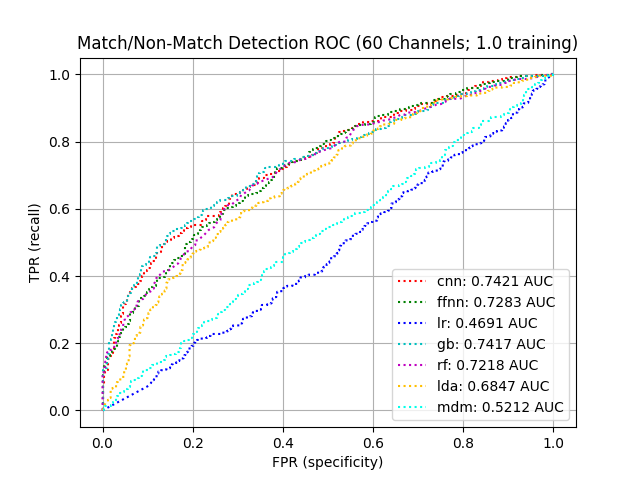}}
    \caption{Scenario 1 benchmarks trained using all 60 channels per trial and 100\% of the trials in the training set.}
\end{figure*}

\subsection{Deep Learning Model Architecture}
\label{sec:dl_model}

We design our deep learning model to be amenable to learning from relatively few EEG trials in comparison to data volumes seen in other domains, e.g., computer vision, NLP, etc. To this end, we introduce a convolutional architecture which explicitly sums responses from feature maps of varying temporal bandwidth (i.e., parallel convolutions), rather than hierarchically stacking and pooling feature maps. The rationale here is to explicitly enforce consideration of salient data at multiple time scales, rather than rely on multi-time scale data propagating through stacked convolutions. While the latter design could be effective, we surmise that it is also more prone to overfitting on EEG-scale datasets. A number of changes/refinements could be integrated, e.g., channel-wise convolutions from \cite{lawhern2018eegnet}, which could be an interesting future study. We note that we do not conduct a thorough architectural exploration in this paper, since the aim of our present study is to explore cross-subject transfer learning rather than perform detailed architectural comparisons.

Specifically, our model takes as input pre-specified channels from one second of EEG data (one trial) at a sampling rate of 256 Hz. This corresponds to 256 input dimensions per-channel. The input data is fed to three separate convolutional layers in parallel with varying kernel sizes of 3, 7, and 11, respectively. Each convolutional layer contains 32-filters and padding is performed so that outputs are equidimensional. The outputs are then summed element-wise to enforce explicit consideration of variable temporal bandwidths during optimization.  Max pooling is then applied, using a kernel size of 2, and the convolutional outputs are then rasterized to a 1D vector of 4096 dimensions. This vector is then passed through a layer normalization layer (LNorm), followed by an exponential linear unit (ELU) activation, then two subsequent stacks of linear layers followed by LNorm and ELUs reduce the dimension first to 1024 then to 128. A final linear layer squashes the output to 1D and a sigmoid link function is applied. 

We trained the network for 7 epochs (the point at which repeated testing showed stabilized loss), using a minibatch size of 128 trials over approximately 3,700+ trials per epoch. For the baseline model, we utilized a cross-entropy loss between the model output and the corresponding (binary) label during backpropagation.

\subsection{Benchmark Comparison Models (Scenario 1)}
\label{sec:benchmark_models}

To justify the application of our convolutional neural network (CNN) architecture, we benchmark performance against a number of more traditional classifiers in Sec. \ref{sec:experiments} on the Scenario 1 test partition. These classifiers include a dense feed-forward neural network (FFNN), logistic regression (LR), a random forest (RF), gradient boosting (GB), linear discriminant analysis (LDA), and Riemannian minimum distance to mean (RMDM). Where applicable, hyperparameters were selected using a grid search over the Scenario 1 validation set.

\subsection{Transfer Learning Comparison Models (Scenario 2)}
 \label{sec:transfer_benchmark}
  For our transfer learning evaluation, we train several variations of our base CNN architecture (cf. \ref{sec:dl_model}) on the Scenario 2 training partition. We then fine-tune on the Scenario 2 fine-tuning partitions (F1, F1+F2, F1+F2+F3), and assess performance on the test partition. Model variations are as follows:

\noindent  
{\bf Baseline: } The CNN's weights were initialized using a Xavier initialization and the CNN was trained over F1, F1+F2, and F1+F2+F3 folds using a binary cross-entropy loss. The T fold was used for evaluation.

\noindent  
{\bf Variation 1: Same-Task Binary Pre-Training; Binary Fine Tuning: } The CNN was trained using the Scenario II pre-training partition, directly on labels, under a binary cross-entropy loss. Fine Tuning was performed over over F1, F1+F2, and F1+F2+F3 at a reduced learning rate. This experiment was repeated, freezing all weights up through the penultimate layer of the network.

\noindent  
{\bf Variation 2: Auto-Encoder Pre-Training; Binary Fine-Tuning: } Similar to Variation 1, except instead of pre-training with a binary cross-entropy loss, an the penultimate layer of the CNN served as the latent space of an auto-encoder. The auto-encoder was trained under a mean squared error (MSE) loss. The fine-tuning procedure was identical to that of Variation 1.

\noindent  
{\bf Variation 3: Siamese Pre-Training; Binary Fine-Tuning: } Similar to Variation 1, except instead of pre-training with a binary cross-entropy loss, a Siamese representation with contrastive loss was employed. The fine-tuning procedure was identical to that of Variation 1.

\noindent  
{\bf Variation 4: Cross-Task Binary Pre-Training; Binary Fine-Tuning: } Similar to Variation 1, except labels from one task were used in pre-training while labels from another task were used for fine-tuning and evaluation.

For Variations 1-3 we conducted our evaluation using both High-Risk/Low-Risk and Match/Non-Match labels. For Variation 4, the cross-task experiment, we used Match/Non-Match labels for pre-training and High-Risk/Low-Risk labels for fine-tuning.

%   \begin{itemize}
%   \item Fine-tuning with reduced learning rate with and without weight freezing on all but the penultimate layer. 
%   \item Cross-task pre-training and fine-tuning.
%   \item Pre-training with an auto-encoder layer and mean squared error (MSE) reconstruction loss attached to the penultimate layer of the base representation.
%   \item Pre-training a Siamese version of the base representation under a contrastive loss.
%   \end{itemize}

\section{Experiments}
\label{sec:experiments}

\subsection{Scenario 1: Performance Benchmark}

\noindent
{\bf Benchmark (All Data, All Channels): } Results for the High-Risk/Low-Risk benchmark are shown in Fig. \ref{fig:alcoholicFull}. Random forests (RF), Riemannian MDM (MDM), and Gradient Boosting (GB) slightly outperformed our CNN model, with the CNN outperforming the feedforward model (FFNN) and dramatically outperforming logistic regression (LR) and linear discriminant analysis (LDA). Results for the Match/Non-Match benchmark are shown in Fig. \ref{fig:matchFull}, where the CNN outperformed all other classifiers. In contrast to the High-Risk/Low-Risk regime, LDA performance increased, while MDM performance deteriorated.

\begin{figure*}[!t]
        \centering
        \subfloat[High-Risk/Low-Risk]{\label{fig:alcoholicPartialChannels}
        \includegraphics[width=0.45\linewidth]{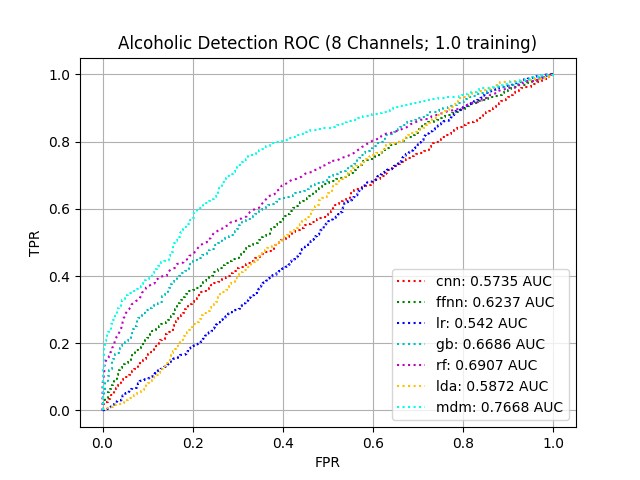}}
        \hfill
        \subfloat[Match/Non-Match]{\label{fig:matchPartialChannels}
        \includegraphics[width=0.45\linewidth]{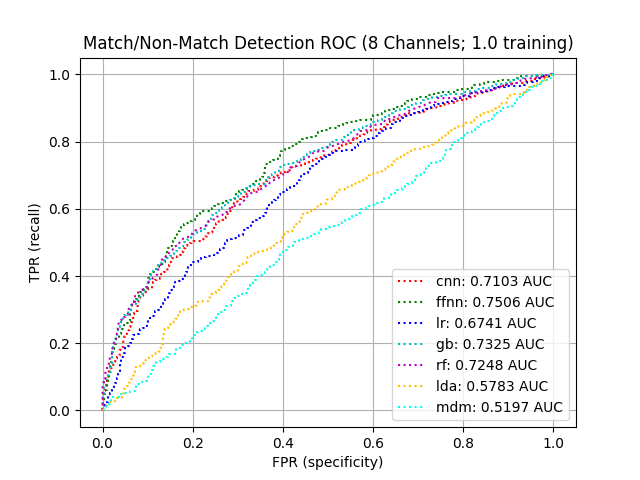}}
    \caption{Scenario 1 benchmarks trained and evaluated using a reduced set of 8 channels and 100\% of trials in the training set.}
\end{figure*}

\noindent
{\bf Effects of Channel Reduction:} In order to assess the effects of number of channels, we sub-selected only eight channels per trial to train and evaluate on. Channels were selected according to standard configuration. Results for High-Risk/Low-Risk detection are shown in Fig.
\protect\ref{fig:alcoholicPartialChannels}. Results for Match/Non-Match detection are shown in Fig. \protect\ref{fig:matchPartialChannels}. Notably, for High-Risk/Low-Risk, areas under Receiver Operating Characteristic (ROC) curves of all classifiers except MDM either remained roughly the same or were reduced; for MDM, performance significantly increased with lower input dimension. For Match/Non-Match, performance of all classifiers remained roughly the same or was reduced, with MDM still performing poorly. However, performance degradation across all classifiers was more graceful than under the High-Risk/Low-Risk regime.

\begin{figure*}[!t]
        \centering
        \subfloat[High-Risk/Low-Risk]{\label{fig:alcoholicPartialData}
        \includegraphics[width=0.45\linewidth]{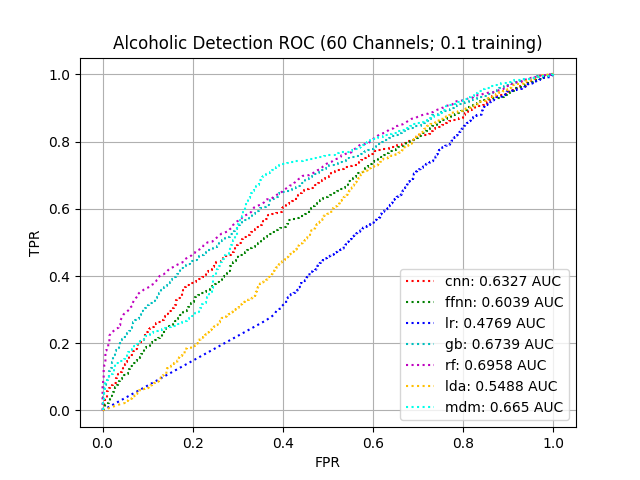}}
        \hfill
        \subfloat[Match/Non-Match]{\label{fig:matchPartialData}
        \includegraphics[width=0.45\linewidth]{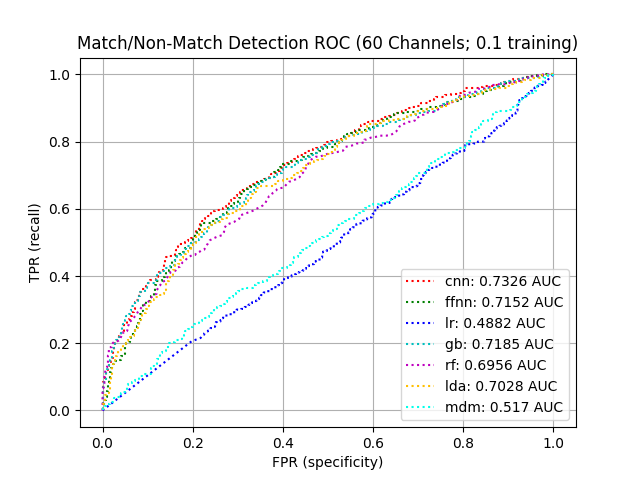}}
    \caption{Scenario 1 benchmarks, trained using 60 channels per trial and 10\% of the trials in the training set.}
\end{figure*}

\noindent
{\bf Effects of Data Reduction:} In order to assess sensitivity of our classifiers to data size,  we re-assessed the Scenario 1 benchmarks, training with only 10\% of the training trials. Trials were sampled according to a uniform random distribution. Results from the High-Risk/Low-Risk and Match/Non-Match experiments under this reduced data regime are shown in Fig. \ref{fig:alcoholicPartialData} and Fig. \ref{fig:matchPartialData} respectively. 

While performance generally deteriorated under both regimes compared with using all trials in train, the random forest was relatively unaffected for High-Risk/Low-Risk detection (net AUC slightly increased). Even with only 10\% of the training set, the CNN remained the top performing model in the Match/Non-Match regime.

\begin{figure*}[!t]
        \centering
        \subfloat[High-Risk/Low-Risk Baseline]{\label{fig:transferAlcoholicRocsBaseline}
        \includegraphics[width=0.45\linewidth]{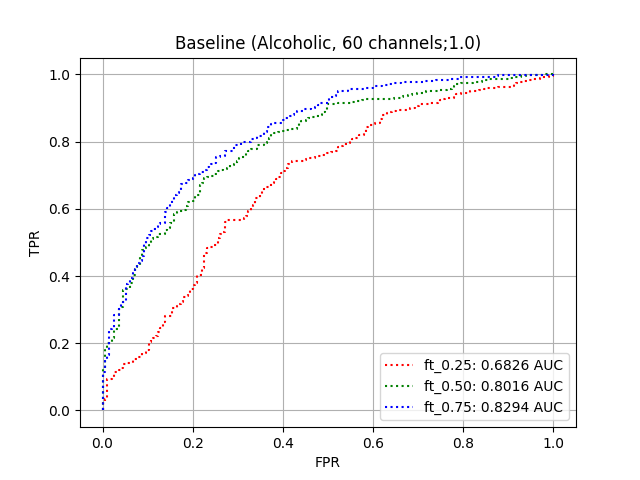}}
        \hfill
        \subfloat[High-Risk/Low-Risk Top Performer (Siamese Pre-Train)]{\label{fig:transferAlcoholicRocsSiamese}
        \includegraphics[width=0.45\linewidth]{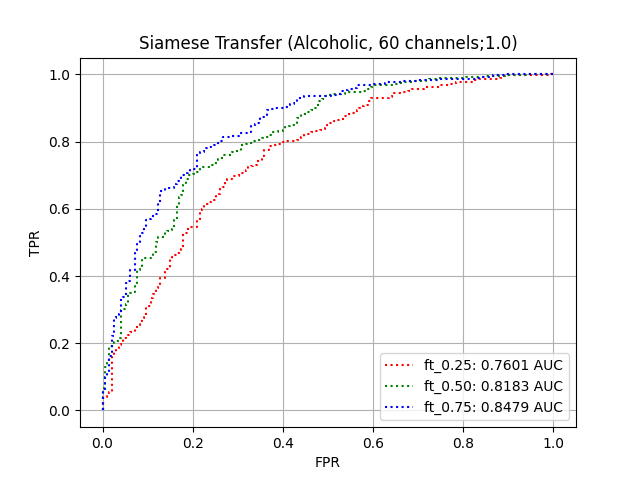}}
    \caption{Scenario 2 High-Risk/Low-Risk benchmarks.}
\end{figure*}

\subsection{Scenario 2: Transfer Learning Evaluation}
% \noindent
% {\bf High-Risk/Low-Risk Regime:} Baseline transfer results for High-Risk/Low-Risk evaluation are shown in Fig. \ref{fig:transferAlcoholicRocsBaseline}. The top performing transfer model was obtained via pre-training under a Siamese regime, then fine-tuning over the entire model at a reduced learning rate. Results from this model are shown in Fih. \ref{fig:transferAlcoholicRocsSiamese}.

% \noindent
% {\bf Match/Non-Match Regime:} Baseline results for the Match/Non-Match transfer evaluation are shown in Fig. \ref{fig:transferMatchRocsBaseline}. The top performing model was obtained by pre-training with a binary cross-entropy loss, freezing all weights up to the penultimate layer, then fine-tuning the unfrozen weights; see Fig. \ref{fig:transferMatchRocsFreeze}.

\begin{table}[!h]
	\caption{AU-ROCs High-Risk/Low-Risk}

	\label{tab:hr_lr_rocs}
	\begin{tabular}{lrrr}%
%		\toprule
		\textbf{Experiment}&\textbf{FT 0.25}&\textbf{FT 0.5}&\textbf{FT 0.75}\\
%		\midrule
		Baseline &0.683&0.802&0.829\\
		\textbf{Siamese (no freeze)}&\textbf{0.760}&\textbf{0.818}&\textbf{0.848}\\
        Siamese (freeze)&0.727&0.806&0.835\\
        Binary (freeze)&0.512&0.526&0.523\\
        Binary (no freeze)&{0.515}&{0.510}&{0.497}\\
        Auto-encoder (freeze)&0.655&0.712&0.749\\
        Auto-encoder (no freeze)&0.638&0.698&0.734\\

	\end{tabular}

\end{table} 

\begin{table}[!h]
	\caption{AU-ROCs Match/No Match}

	\label{tab:match_nonmatch_rocs}
	\begin{tabular}{lrrr}%
%		\toprule
		\textbf{Experiment}&\textbf{FT 0.25}&\textbf{FT 0.5}&\textbf{FT 0.75}\\
%		\midrule
		Baseline &0.608&0.712&0.659\\
		Siamese (no freeze)&0.660&0.703&0.771\\
        Siamese (freeze)&0.635&0.717&0.773\\
        \textbf{Binary (freeze)}&\textbf{0.787}&\textbf{0.796}&\textbf{0.801}\\
        {Binary (no freeze)}&{0.765}&{0.773}&{0.784}\\
        Auto-encoder (freeze)&0.568&0.642&0.673\\
        Auto-encoder (no freeze)&0.543&0.631&0.680\\
        
	\end{tabular}

\end{table}

Transfer results for the High-Risk/Low-Risk evaluation are shown in Table \ref{tab:hr_lr_rocs} in terms of area under each ROC (AU-ROC). The full ROC curves for the baseline model and top performing model are shown in Figs. \ref{fig:transferAlcoholicRocsBaseline} and \ref{fig:transferAlcoholicRocsSiamese} respectively. The top performing transfer model was obtained via pre-training under a Siamese regime, then fine-tuning over the entire model at a reduced learning rate. 

\begin{figure*}[!b]
        \centering
        \subfloat[Match/Non-Match Baseline]{\label{fig:transferMatchRocsBaseline}
        \includegraphics[width=0.45\linewidth]{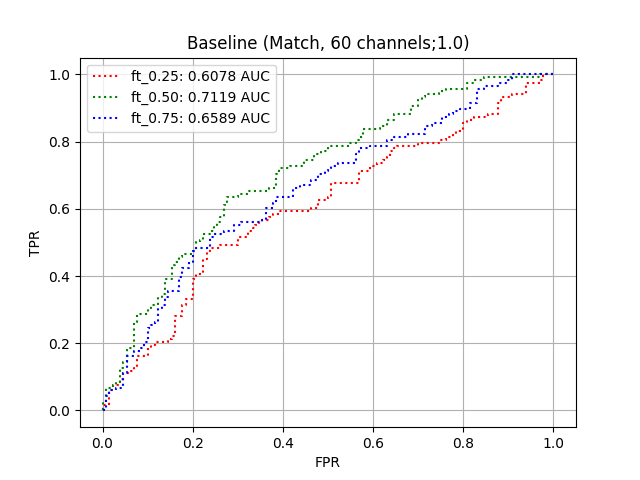}}
        \hfill
        \subfloat[Match/Non-Match Top Performer (Frozen Weights from Pre-Training Through Penultimate Layer)]{\label{fig:transferMatchRocsFreeze}
        \includegraphics[width=0.45\linewidth]{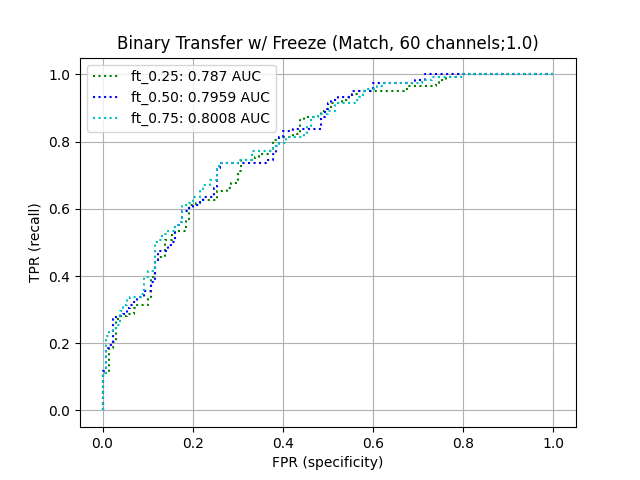}}
    \caption{Scenario 2 Match/Non-Match benchmarks.}
\end{figure*}

Transfer results for the Match/Non-Match evaluation are shown in Table \ref{tab:match_nonmatch_rocs} in terms of area under each ROC curve. The full ROC curves for the baseline model and top performing model are shown in Figs. \ref{fig:transferMatchRocsBaseline} and \ref{fig:transferMatchRocsFreeze} respectively. The top performing model was obtained by pre-training with a binary cross-entropy loss, freezing all weights up to the penultimate layer, then fine-tuning the unfrozen weights.

Transfer results for the Cross-Task transfer evaluation are shown in Table \ref{tab:match_nonmatch_rocs}. ROCs for these experiments are not shown as these models significantly under-performed the baseline.

\begin{table}
	\caption{AU-ROCs Cross-Task}

	\label{tab:cross-task_rocs}
	\begin{tabular}{lrrr}%
%		\toprule
		\textbf{Experiment}&\textbf{FT 0.25}&\textbf{FT 0.5}&\textbf{FT 0.75}\\
%		\midrule
        Binary cross-task (freeze)&0.581&0.597&0.617\\
        Binary cross-task (no freeze)&0.617&0.634&0.665\\
        
	\end{tabular}

\end{table}
\section{Discussion and Conclusions}
\label{sec:conclusion}

We have trained a deep learning CNN architecture across multiple heterogeneous subjects, and shown that cross-subject training  works effectively across two different tasks: detection of risk for alcoholism (High-Risk/Low-Risk) and detection of stimulus (Match/Non-Match). For the Match/Non-Match detection, the CNN outperforms all traditional benchmarks.

For High-Risk/Low-Risk detection, the CNN under-performs ensemble methods and Riemannian MDM classifier in terms of raw AU-ROC, but over-performs other benchmarks. In lower false positive rate (FPR) ranges, both the CNN and ensemble methods outperform Riemannian MDM. Note that the instability in the MDM ROC progressed as we increased the number of channels during our benchmark testing. Moreover, in contrast to the CNN, which seems to benefit from additional training data and number of electrodes, the Riemannian classifier performed better with fewer number of electrodes and its performance advantage over other classifiers was more pronounced in the low-data regime. This suggests that the CNN may scale more gracefully with increased data volumes than a Riemannian classifier, which is consistent with findings in the EEG literature \cite{lotte2018review,rodrigues2019exploring} wherein Riemannian classifiers perform well with small numbers of electrodes and do not scale well with larger input sizes. 

Most significantly, our transfer learning experiments demonstrate that by starting with a base representation trained across a multitude of heterogeneous subjects and transferring this representation to novel subjects, one can substantially reduce data requirements and/or enhance performance, even when the transfer representation is pre-trained on small data volumes.

For the High-Risk/Low-Risk transfer benchmark, we found that pre-training using a contrastive loss under a Siamese network regime, then fine-tuning over all weights using a reduced learning rate yielded the best transfer performance, while for the Match/Non-Match transfer benchmark, we found that freezing all weights up through the penultimate layer of a pre-trained model, trained on a binary cross-entropy loss, yielded best transfer performance. For the High-Risk/Low-Risk regime, the lack of transfer of the models that were pre-trained using a binary cross-entropy loss was unexpected. The lack of transfer under our cross-task training regime was additionally unexpected. We intend to examine cross-task transfer in more detail in future work.

These findings from our research may lead to benefits in consumer/clinical BCI applications, specifically as dataset sizes grow. With large datasets and consistent data collection protocols, the burden of data collection for training on a specific end-user could be dramatically reduced if not entirely eliminated for certain applications. Logical future directions of research include further exploration of cross-task transfer learning, pre-training across multiple optimization objectives, and refinements/improvements to selected deep transfer architectures.
\section{Acknowledgement}
This work was supported in part by AFWerx STTR award numbers FA864919PA077 and FA864920P0420.

\clearpage
\bibliographystyle{IEEEtran}
\bibliography{references}

%rebuttal removed for camera-ready submission.
%\input{IEEEconf_ICPR2022/sections/rebuttal}

\end{document}